\documentclass[10pt,twocolumn,letterpaper]{article}

\usepackage{iccv}
\usepackage{times}
\usepackage{epsfig}
\usepackage{graphicx}
\usepackage{amsmath}
\usepackage{amssymb}
\usepackage[normalem]{ulem}
\usepackage{color}
\usepackage{comment}
\usepackage{ifthen}

\definecolor{purple}{rgb}{0.6, 0, 0.6}

\newcommand{\figref}[1]{{Figure \ref{fig:#1}}}
\newcommand{\secref}[1]{{Section \ref{sec:#1}}}
\newcommand{\eqnref}[1]{{Eq.\ (\ref{eq:#1})}}
\newcommand{\tabref}[1]{{Table \ref{tab:#1}}}

\newboolean{revising}
\setboolean{revising}{false}
\ifthenelse{\boolean{revising}}
{
	\newcommand{\Hao}[1]{\textcolor{purple}{#1}}
} {
	\newcommand{\Hao}[1]{#1}
}

\usepackage[pagebackref=true,breaklinks=true,letterpaper=true,colorlinks,bookmarks=false]{hyperref}

\iccvfinalcopy 

\def\iccvPaperID{177} 

\ificcvfinal\fi
\begin{document}

\title{Visual Forecasting by Imitating Dynamics in Natural Sequences}

\author{Kuo-Hao Zeng$^{\dagger}$$^{\ddagger}$\enspace 	
William B. Shen$^{\dagger}$\enspace De-An Huang$^{\dagger}$\enspace Min Sun$^{\ddagger}$\enspace Juan Carlos Niebles$^{\dagger}$\\
$^{\dagger}$Stanford University\enspace$^{\ddagger}$National Tsing Hua University\\
\{khzeng, bshen88, dahuang, jniebles\}@cs.stanford.edu
\enspace sunmin@ee.nthu.edu.tw
}

\maketitle
\thispagestyle{empty}

\begin{abstract}

%
\Hao{
We introduce a general framework for visual forecasting, which directly imitates visual sequences without additional supervision.
} As a result, our model can be applied at several semantic levels and does not require any domain knowledge or handcrafted features. We achieve this by formulating visual forecasting as an inverse reinforcement learning (IRL) problem, and directly imitate the dynamics in natural sequences from their raw pixel values. The key challenge is the high-dimensional and continuous state-action space that prohibits the application of previous IRL algorithms. 
We address this computational bottleneck by extending recent progress in model-free imitation with trainable deep feature representations, which
(1) bypasses the exhaustive state-action pair visits in dynamic programming by using a dual formulation and (2) avoids explicit state sampling at gradient computation using a deep feature reparametrization. This allows us to apply IRL at scale and directly imitate the dynamics in high-dimensional continuous visual sequences from the raw pixel values. We evaluate our approach at three different level-of-abstraction, from low level pixels to higher level semantics: future frame generation, action anticipation, visual story forecasting. At all levels, our approach outperforms existing methods.



\end{abstract}

\vspace{-5mm}
\section{Introduction}

Our goal is to expand the boundaries of video analysis towards visual forecasting by building intelligent systems that are capable of imitating the dynamics in natural visual sequences. The idea of visual forecasting has been studied in several contexts, such as trajectory forecasting~\cite{alahi2016social,kitani2012activity,ptf_cvpr2014}, activity prediction~\cite{lan2014hierarchical,soran2015generating}, and future frame synthesis~\cite{kalchbrenner2016video,xue2016visual}. However, most previous work either assumes short forecasting horizons \cite{vondrick2016anticipating,of_iccv2015}, or focuses on a specific application, such as walking trajectories~\cite{alahi2016social,kitani2012activity,xie2013inferring} or human actions~\cite{huang2014action,lan2014hierarchical}, where handcrafted features play an important role. 


\begin{figure}[t]
\begin{center}
  \includegraphics[width=1\linewidth]{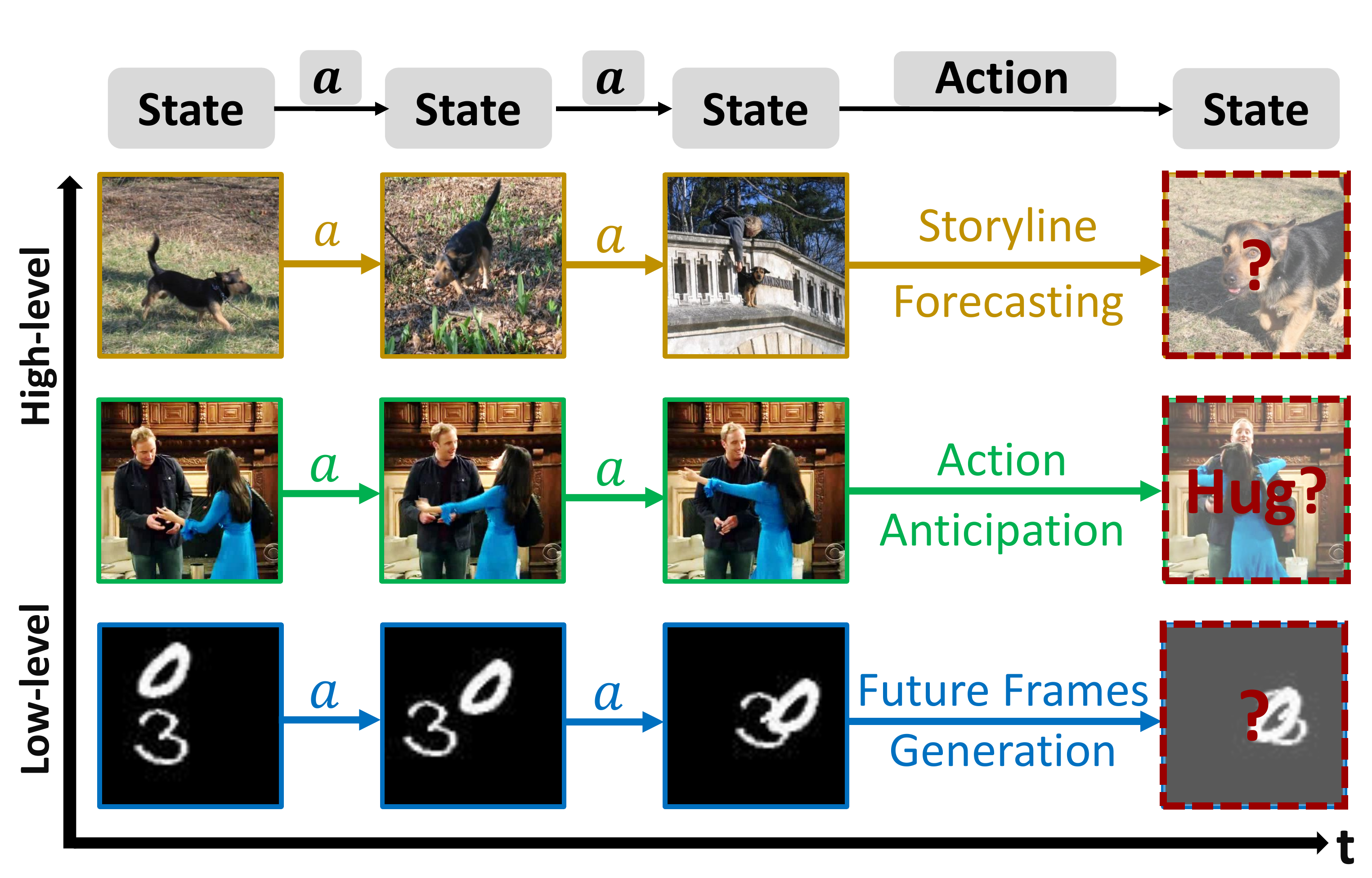}
\end{center}\vspace{-5mm}
   \caption{\small We propose a general framework for visual forecasting. We formulate this as an imitation learning problem, where the natural visual sequences are served as expert behaviors, and visual forecasting is reformulated as learning a policy to reproduce the expert's behavior. Our framework imitate dynamics in visual sequences directly from its raw pixel value and is applicable to tasks with different levels of abstraction.}
   \label{fig:1}
   \vspace{-5mm}
\end{figure}

In this paper, we propose a general framework for visual forecasting\footnote{\url{http://ai.stanford.edu/~khzeng/projects/vfid/}}, where it can be applied to learn a model at any given semantic-level (see \figref{1}) and does not require any domain knowledge or handcrafted features.
Such framework for forecasting can be widely useful in real world scenarios: robot motion planning \cite{rnn-brain4cars-saxena-2016,koppula2015_anticipatingactivities} or warning systems for autonomous vehicles \cite{ACCV_accident}. We achieve this by formulating visual forecasting as an inverse reinforcement learning (IRL) \cite{ng2000algorithms}  problem, where the natural sequences serve as expert demonstrations, and the goal becomes recovering the underlying cost function that prioritizes the entire expert trajectories over others. This eliminates the compounding error problem of single-timestep decisions~\cite{ho2016generative}, and has been proven effective in long-term prediction in both robotics \cite{finn2016guided,kretzschmar2016socially,ziebart2008maximum} and vision \cite{kitani2012activity,mathe2013action}.

Our key insight is to extend previous IRL work in vision to directly imitate the dynamics in visual sequences from their raw pixel values, where no domain knowledge or handcrafted feature would be required. This is in contrast to previous work that applies IRL on handcrafted features, such as the semantic scene labeling for trajectory forecasting~\cite{kitani2012activity,ma2016game} and interaction statistics for dual agent interaction~\cite{huang2014action}.
The main challenge is that both the state and action space would become \emph{high-dimensional} and \emph{continuous} in such setting, which prevent scaling up existing IRL algorithms used in previous work~\cite{ho2016generative,ziebart2008maximum}.



We address this challenge by extending recent progress in model-free imitation learning~\cite{ho2016generative,ho2016model} with the capacity of jointly trainable deep feature representations. More specifically, we extend the Generative Adversarial Imitation Learning framework of Ho and Ermon~\cite{ho2016generative} with deep feature reparametrization to forecast the dynamics in natural visual sequences.
\Hao{The deep representation is fully differentiable and we optimize it jointly using policy gradient, which results in learning an intermediate representation that is best-suited for visual forecasting.}
The resulting model (1) bypasses the dynamic programming based subroutine of~\cite{ziebart2008maximum} that exhaustively visits every state-action pair by using a dual formulation, and (2) avoids explicit state sampling (image synthesis in this case) of gradient computation in~\cite{ho2016generative} with reparametrization using deep feature representations. This allows us to operate beyond low-dimensional state-action spaces and directly forecast the natural sequence from the pixel-level.







We demonstrate the effectiveness of our approach by applying our model to three tasks that require different levels of understanding of visual dynamics.  First, we examine the ability of our method to perform generation of future frames on the moving MNIST data sequences \cite{srivastava2015unsupervised}. Results from future frame generation demonstrate that our framework can capture the low-level (pixel-level) visual dynamics between consecutive frames. Second, we examine the ability of our method to perform action anticipation by unsupervised training on the THUMOS15 dataset \cite{THUMOS15}, and evaluating the resulting model on the TV Human Interactions dataset \cite{patron2010high}. Results from action anticipation show that our framework is effective in modeling mid-level (motion-level) visual dynamics in natural videos. Finally, we apply our method on the Visual Storytelling dataset \cite{huang2016visual} and try to forecast the next photo in a story composed of images. Results on this task demonstrate that our framework can generalize to visual dynamics on a highly abstract level (semantic-level). In each task, our method outperforms all existing methods.


In summary, the main contributions of our work are: (1) We propose a general IRL based formulation for visual sequence forecasting that requires no domain knowledge or handcrafted features. (2) We address the computational bottlenecks of IRL when operating at high-dimensional continuous spaces by using a dual formulation and reparametrization with deep feature representations. (3) We evaluate our method on three tasks at different levels of abstraction and show that our framework outperforms all existing methods.


\section{Related Work}\label{RW}



\noindent \textbf{Visual Prediction.}
There has been growing interest in developing computational models of human activities
that can extrapolate unseen information and predict future unobserved activities \cite{finn2016unsupervised,mathieu2015deep,park2016egocentric,srivastava2015unsupervised,vondrick2016generating,vae_eccv2016,xue2016visual,zeng2017_risk_assessment}. Some of the existing approaches \cite{finn2016unsupervised,mathieu2015deep,vondrick2016generating,vondrickgenerating,vae_eccv2016,xue2016visual} tried to generate realistic future frames using generative adversarial networks \cite{goodfellow2014generative}. Unlike these methods, we emphasize longer-term sequential dynamics in videos using inverse reinforcement learning. Other line of work attempted to infer the action or human trajectories that will occur in the subsequent time-step based on previous observation \cite{alahi2016social,huang2014action,kitani2012activity,lan2014hierarchical,vondrick2016anticipating}. Our model directly imitates the natural sequence from the pixel-level and assumes no domain knowledge.


\noindent \textbf{Reinforcement Learning in Computer Vision.}
Reinforcement learning (RL) achieves remarkable success in multiple domains ranging from  robotics \cite{gu2016deep}, computer vision \cite{butko2009optimal,mnih2014recurrent,paletta2005q} and natural language processing \cite{li2017adversarial,yu2016seqgan}. In the RL setting, the reward function that the agent aims to maximize is given as signal for training, where the goal is to learn a behavior that maximizes the expected reward. On the other hand, we work on the inverse reinforcement learning (IRL) problem, where the reward function must be discovered from demonstrated behavior~\cite{ho2016generative,ng2000algorithms,ziebart2008maximum}. This is inspired by recent progress of IRL in computer vision~\cite{finn2016guided,huang2014action,kitani2012activity,ma2016game,mathe2013action,ptf_cvpr2014}. Nonetheless, these frameworks require heavy use of domain knowledge to construct the handcrafted features that are important to the task. Unlike these approaches, we aim to generalize IRL to natural sequential data without annotations. 


\noindent \textbf{Unsupervised Representation Learning in Video.} Our combination of deep neural networks with IRL to directly imitate natural videos is related to recent progress in unsupervised representation learning in videos. The temporal information in the video has been utilized to learn useful features based on tracking~\cite{wang2015unsupervised}, flow prediction~\cite{of_iccv2015}, future frame synthesis~\cite{srivastava2015unsupervised}, frame order~\cite{misra2016shuffle}, and object movement~\cite{pathak2016learning}. Our work utilizes not only the frame-to-frame dynamics, but tries to imitate the entire sequence directly to utilize long-range temporal information.

\noindent\textbf{Generative Adversarial Learning.} Our extension of generative adversarial imitation learning~\cite{ho2016generative} is related to recent progress in generative adversarial networks (GAN)~\cite{goodfellow2014generative}. While there has been multiple works on applying GAN to image and video~\cite{pix2pix2016,radford2016unsupervised,vondrick2016generating,yan2016attribute2image}, we extend it to long-term prediction of natural visual sequences and directly imitate the high-dimensional continuous sequence.

\begin{figure}
\begin{center}
\includegraphics[width=1\linewidth]{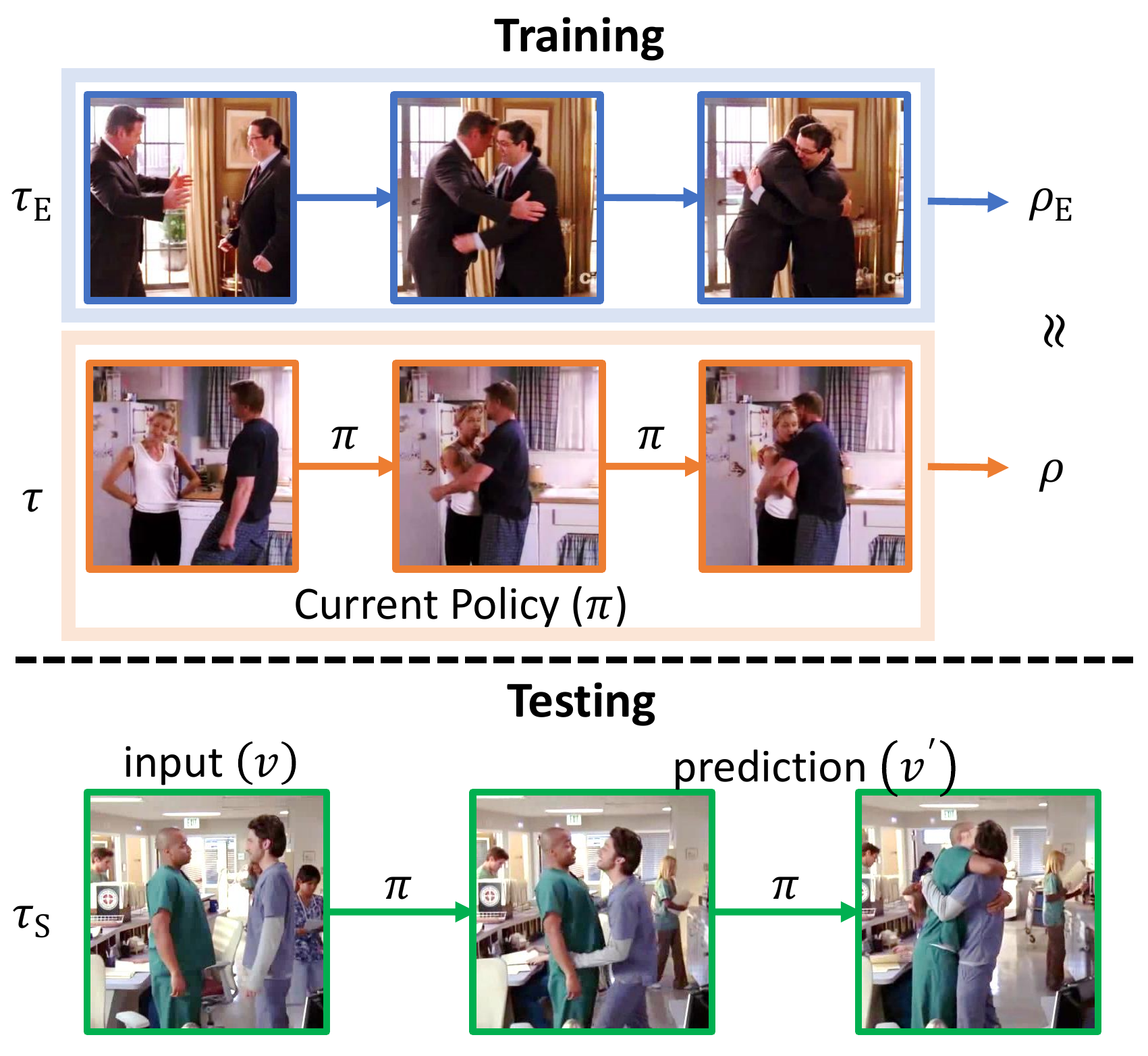}
\end{center}\vspace{-4mm}
   \caption{\small
   We formulate visual prediction as an imitation learning problem, where the goal becomes learning a policy $\pi$ that can generate trajectory $\tau$ following the behavior of natural visual sequence, or expert demonstration $\tau_E$. This is reformulated as finding a policy $\pi$ that has a state-action pair visit distribution $\rho$ that is similar to that of the expert ($\rho_E$). At test time, we thus have a learned policy $\pi$ that can mimic the behavior of natural visual sequence and perform visual prediction.}
   \vspace{-5mm}
   \label{fig:system}
\end{figure}

\section{Forecasting Natural Sequences}
\label{sec:method}

Our goal is to build a system that can forecast the dynamics of natural videos. We formulate this as an inverse reinforcement learning problem, where the goal is to recover the underlying cost function that induces the natural visual sequence. This is equivalent to learning a policy $\pi(v'|v)$ that parameterizes the transition from a current frame $v$ to the next frame $v'$. As shown in \figref{system}, at training time, our model observes and imitates a large amount of natural visual sequences to learn the policy. At testing time, the learned policy $\pi(v'|v)$ can then be directly applied to an image for long-term visual prediction.
Despite its many potential applications, learning $\pi(v'|v)$ is challenging as $v$ is an image that is high-dimensional and continuous. This makes both the exhaustive state-action pairs visit and the explicit state sampling in previous work~\cite{ho2016generative,ziebart2008maximum} intractable at scale. Our solution to this extends and reparametrizes the model from Ho and Ermon~\cite{ho2016generative} with deep representations. Our framework bypasses exhaustive state-action pair visit of~\cite{ziebart2008maximum} by the dual formulation of~\cite{ho2016generative}, and avoids explicit state sampling (which is equivalent to frame synthesis) during the gradient computation in~\cite{ho2016generative}.

In the following, we discuss how we formulate natural visual sequences as Markov Decision Processes so that our inverse reinforcement learning framework can be applied.

\subsection{Visual Sequences as Markov Decision Process}
\label{sec:mdp}

In this work, we model natural visual sequences as Markov Decision Processes (MDP) \cite{puterman2014markov}. At each time step, the sequence is at some frame/image $v$, which is defined as the \emph{state} of our MDP. Given a chosen action $a$ from the state, the process responds by moving to a new state, or frame $v'$ based on the transition model $p(v'|v,a)$ and receives a cost $c(v,a)$. In this work, the transition model is \emph{deterministic} as we are directly imitating the natural visual sequences. Therefore, $p(v'|v, a) = 1$  for the  single visual state $v' = v_a$ that we can visit after taking action $a$, and is zero otherwise. \Hao{In this case, the policy $\pi(v_a | v)$ of moving to a state $v_a$ is equivalent to choosing the corresponding action $\pi(a | v)$.} As result, we are able to use $\pi(v'|v)$ and $\pi(a|v)$ interchangeably depending on the context (same for $c(v,v')$ and $c(v, a)$). Our goal of inverse reinforcement learning (IRL) is in contrast to reinforcement learning (RL), where the cost function is observed and the goal is to learn the optimal policy $\pi(a|v)$ that minimizes the expected cost. In IRL, the cost function is \emph{not} observed, and the goal is to imitate the behaviors of the expert by recovering the underlying cost function $c(v,a)$ that induces the observed optimal behavior by the expert.





\begin{figure*}[t]
\begin{center}
  \includegraphics[width=1\linewidth]{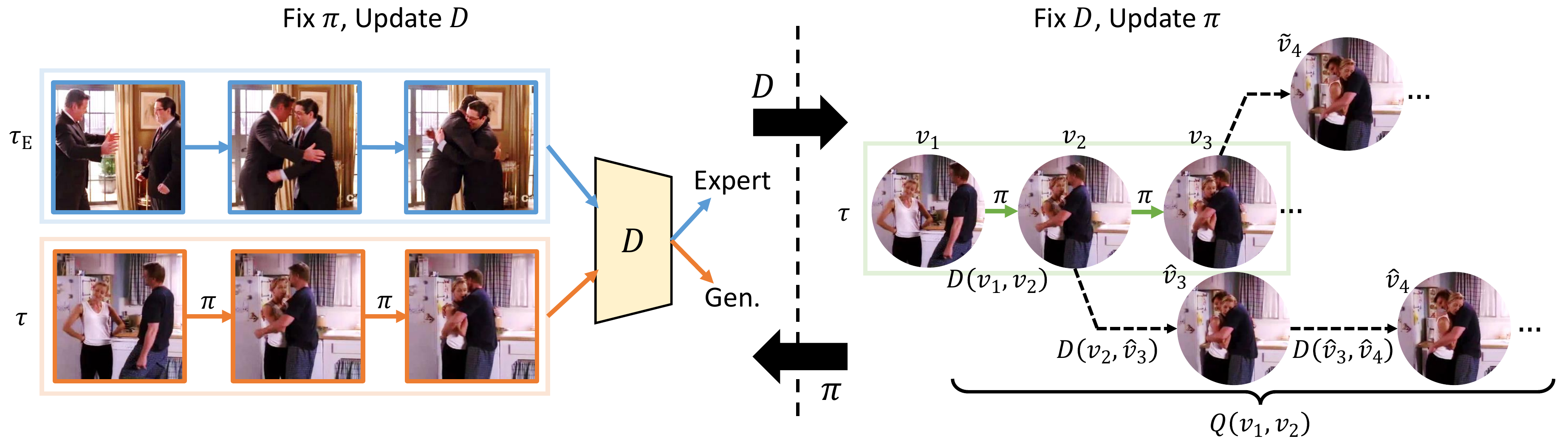}
\end{center}\vspace{-4mm}
   \caption{\small Overview of our imitation learning. The optimization iterates between policy update and cost function update. During the cost ($D$) update, the goal is to find a cost that best differentiate expert's trajectory $\tau_E$ from our generated trajectory $\tau$. During the policy ($\pi$) update, we estimate the expected cost $Q$ and use policy gradient with roll out to minimize it (see more detail in supplementary). This framework also has a generative adversarial learning interpretation, where the cost is operating as the discriminator and the policy is functioning as the generator.}
   \label{fig:learning}
   \vspace{-5mm}
\end{figure*}

\subsection{Imitating Natural Visual Sequences}

We formulate the visual prediction as an inverse reinforcement learning problem, where the goal is to imitate the behavior of natural sequences that are treated as expert demonstrations. In this section, we discuss the model-free imitation learning framework of Ho and Ermon~\cite{ho2016generative} that allows us to bypass the exhaustive state-action pair visit in dynamic programming of Maximum Entropy (MaxEnt) IOC~\cite{ziebart2008maximum} used in previous works~\cite{huang2014action,kitani2012activity}. In the next section, we discuss how we apply it efficiently to natural visual sequences from the pixel level.




 In contrast to reinforcement learning, the cost function in IRL is not given and has to be recovered from expert demonstrations $\tau_E$. We follow previous work and adopt the maximum causal entropy framework~\cite{ziebart2008maximum} for the optimization of our cost function:
\begin{equation}
\label{eq:maxent_irl}
    \max_{c\in C} \, ( \, \min_{\pi\in \Pi} -H(\pi) + \mathbb{E}_{\pi}[c(v,a)] \, ) - \mathbb{E}_{\pi_{E}}[c(v,a)],
\end{equation}
where $\Pi$ is a policy space and $H(\pi)\triangleq\mathbb{E}_{\pi}[-log \, \pi(a|v)]$ is the $\gamma$-discounted causal entropy \cite{bloem2014infinite} of the policy $\pi$.

The standard approach for this IRL problem used in previous work~\cite{kitani2012activity,ziebart2008maximum} involves a RL inner loop
that utilizes dynamic programming to exhaustively visit every state and the available actions for value function computation. However, such procedure is intractable in our problem, as both our state and action spaces are high-dimensional and continuous. The key observation to resolve this problem is that IRL \eqnref{maxent_irl} is actually a dual of an occupancy measure matching problem~\cite{ho2016model}. Here, the occupancy measure $\rho_\pi(v,a)$ of a policy $\pi$ is defined as the distribution of the visit of state-action pairs while following the policy $\pi$. This reformulates \eqnref{maxent_irl} as:
\begin{equation}
\label{eq:occu_match}
\min_{\pi} d_\psi (\rho_\pi, \rho_{\pi_E}) - H(\pi),
\end{equation}
where $d_\psi(\cdot,\cdot)$ measures the difference between two occupancy measures depending on the regularizer $\psi$. Intuitively, this aims to induce a policy that visits state-action pairs with similar frequency to that of the expert policy. This allows us to use the demonstrated state-action pairs of expert directly as learning signal to encourage the policy toward expert-like region in state-action space and bypass the exhaustive visit of dynamic programming. 

As presented in~\cite{ho2016generative}, under certain selection regularizer $\psi$, $d_\psi(\cdot,\cdot)$ becomes:
\begin{equation}
\label{eq:D}
\max_D \mathbb{E_{\pi_{\theta}}}[\log(D(v,a))] + \mathbb{E_{\pi_{E}}}[\log(1-D(v,a))],
\end{equation}
where $D(v,a)$ can be interpreted as a discriminator that tries to distinguish the state-action pairs from the expert or the induced policy. This interpretation connects imitation learning with generative adversarial network (GAN)~\cite{goodfellow2014generative}. Essentially, our imitation learning of natural visual sequences can be seen as jointly training (1) a discriminator that tries to distinguish our generated sequence from expert's and (2) a policy that forms as a generator that uses the learning signal provided by the discriminator to move toward expert-like regions. We refer the readers to~\cite{ho2016generative} for more details of the algorithm. The important note is that no specific restriction would be put on the form of approximation for both $D$ and $\pi$ and this allows us to directly imitate high-dimensional continuous sequence.

\noindent\textbf{Optimization.}
Let $D_w$ be $D$ parameterized by weights $w$, $\pi_\theta$ be $\pi$  parameterized by $\theta$, $\tau = \{v_t\}$ be trajectory of states $v^t$ from current policy, and $\tau_E = \{v^E_t\}$ be expert trajectory given as demonstration, the optimization of \eqnref{occu_match} with \eqnref{D} alternates between the maximizing the discriminative power of $D_w$ through (the left panel in \figref{learning}):
\begin{align}
    &\mathbb{E_{\tau\sim\pi_{\theta}}}[\nabla_{w} \log(D_{w}(v_t,v_{t+1}))] \\&+ \mathbb{E_{\tau_{E}\sim\pi_{E}}}[\nabla_{w} \log(1-D_{w}(v_t^E,v_{t+1}^E))],
    \label{eq:delta_D}
\end{align}
and a policy gradient step (the right panel in \figref{learning}):
\begin{equation}
  \mathbb{E_{\tau\sim\pi_{\theta}}}[\nabla_{\theta}\log(\pi_{\theta}(v_{t+1}|v_t))Q(v_t,v_{t+1})]
  \label{eq:PG}
\end{equation}
that aims to maximize the expected return $Q(v_t, v_{t+1})$, which is in turn defined as 
\begin{equation}
    Q(v,v') = \mathbb{E_{\tau\sim\pi_{\theta}}}[\log(D_{w}(v_t,v_{t+1}))|v_{0}=v,v_{1}=v'],
    \label{eq:Q}
\end{equation}
the expectation of confusing the discriminator that the generated sequence $\tau$ is a expert's sequence.

\subsection{Efficient Sampling with Deep Representation}
\label{sec:CNN_IRL}


Now we have formulated visual prediction as an IRL problem and presented a generative adversarial imitation learning framework~\cite{ho2016generative} that is applicable to our problem, because it bypasses the dynamic programming based subroutine that iterates through all the states and actions. However, we would like to point out there is still a remaining computational bottleneck that prevents the direct application of this framework to the image space.

Notice in the gradient computation for both the discriminator and the policy (\eqnref{delta_D} and \eqnref{PG}) that we are required to sample trajectories $\tau$ from the current policy $\pi_\theta$. This is equivalent to sampling a sequence of images/frames at each gradient computation. The problem of image synthesis itself is challenging and computational intensive~\cite{kalchbrenner2016video}, and prevents the direct application of their framework to large scale visual sequence datasets.

\begin{figure}[t]
\begin{center}
  \includegraphics[width=1\linewidth]{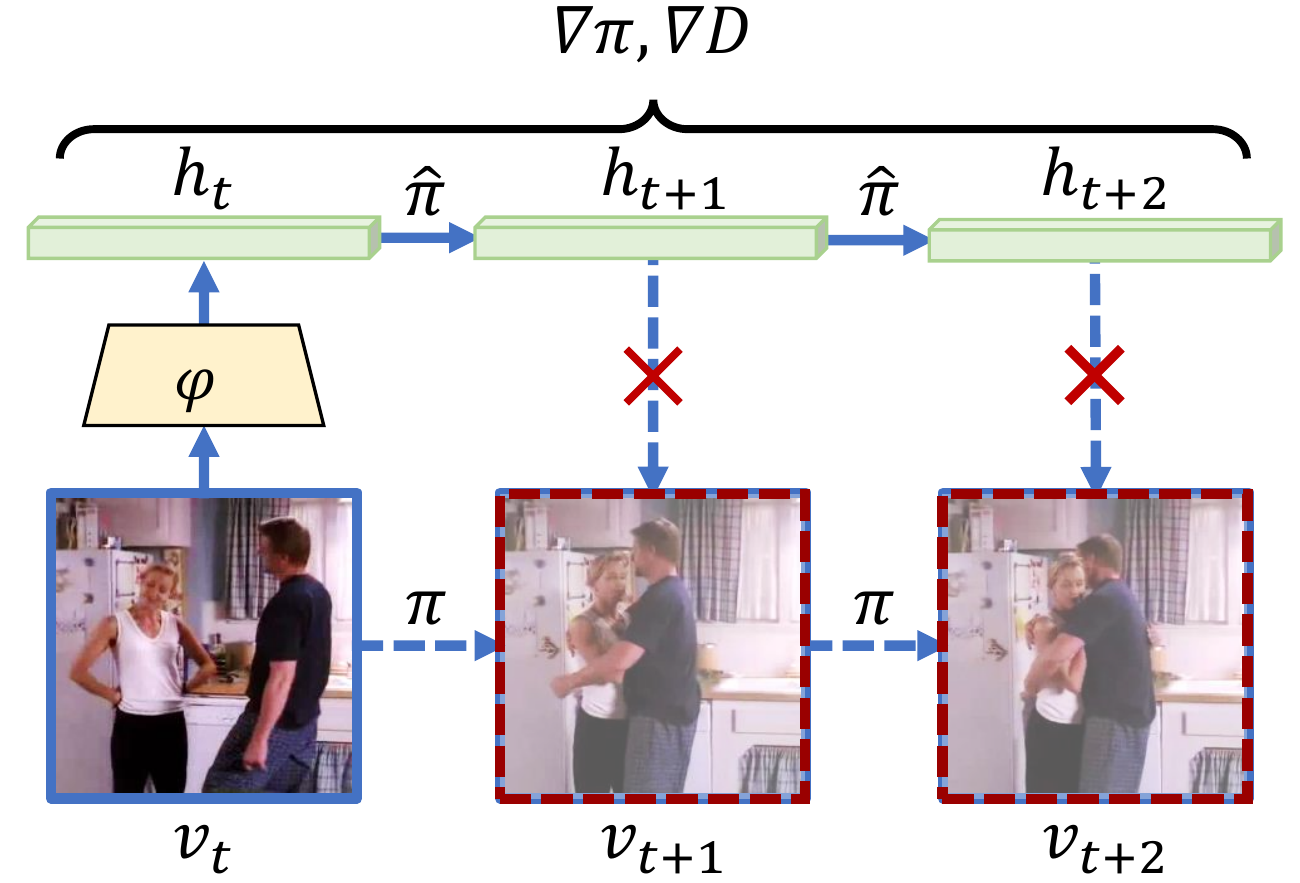}\vspace{-4mm}
\end{center}
   \caption{\small We reparametrize the policy with deep feature representation to avoid explicit state sampling during the gradient computation. The steps of sampling from $\pi$ is decomposed in to (1) applying $\phi(v_t)$ to get the hidden representation $h_t$, (2) sampling from reparametrized policy $\hat{\pi}$ to get $h_{t+1}$, (3) applying $\phi^{-1}(h_{t+1})$ for the next state. However, the third step is not required as we subsequently apply $\phi(\phi^{-1}(h_{t+1}))$ for the sampling in the next time step and cancel out the inverse operation.}
   \label{fig:sampling}
   \vspace{-5mm}
\end{figure}

Our key insight to address this is to parameterize both the discriminator $D_w$ and the policy $\pi_\theta$ with a intermediate representation $\phi(\cdot)$ that is feasible to sample at the optimization, but still captures sufficient information from the image:
\begin{equation}
\pi(v_{t+1}| v_{t}) = \hat{\pi}(\phi(v_{t+1})| \phi(v_{t})),
\end{equation}
\begin{equation}
D(v_t, v_{t+1}) = \hat{D}(\phi(v_t), \phi(v_{t+1})).
\end{equation}
In this case, the sampling of $v_{t+1}$ from $v_t$ is a three step process (all steps in \figref{sampling}):
\begin{align}
v_{t+1} = \phi^{-1}(h_{t+1}),\quad h_{t+1} \sim \hat{\pi}_\theta(\cdot|h_t),\quad h_t = \phi(v_t).
\end{align}
The main advantage of this parameterization is that the challenging $\phi^{-1}$ (image synthesis from hidden representation) is actually not required in the gradient computation, since:
\begin{align}
&\pi(v_{t+1}| v_{t}) = \hat{\pi}(\phi(v_{t+1})| \phi(v_{t})) \\
&= \hat{\pi}(\phi(\phi^{-1}(h_{t+1}))|h_t) = \hat{\pi}(h_{t+1}|h_t),
\end{align}
and similar for the discriminator. In this case, only the intermediate representation is required to be sampled for the gradient computation (the top path in \figref{sampling}). In this work, we use the deep convolutional neural networks as our $\phi(\cdot)$ because of its success in representing the visual information~\cite{krizhevsky2012imagenet}. It is important to note that this $\phi(\cdot)$ is also fully differentiable and we approximate it by optimizing it jointly in the policy gradient step of \eqnref{PG}. This reparametrization allows us to directly imitate natural visual sequences from its pixel value and goes beyond the computational limitation in  previous works.




\section{Experiments}\label{Exp}

We evaluate our model on three challenging tasks at three different levels of abstraction. The first is future frame generation in the moving MNIST dataset~\cite{srivastava2015unsupervised}, where the image synthesis ($\phi^{-1}$) is feasible to verify the effectiveness of our imitation learning formulation. The second task is action anticipation on the TV Human Interactions dataset~\cite{patron2010high}. We verify the predictive capacity of our model directly imitating natural video~\cite{THUMOS15} from the pixel-level with our deep reparametrization. Finally, we apply our method on the Visual Storytelling dataset~\cite{huang2016visual} to predict the next photo in a storyline that consists of five photos. This is extremely challenging as there can be no clue from direct visual appearance matching. We show that our method can still mimic human behavior at the \emph{semantic}-level from the pixel values.



\subsection{Future Frame Generation}\label{sec:FFG}
In this experiment, we examine our method's ability to perform future frame generation by training on the moving MNIST data sequences \cite{srivastava2015unsupervised}. Unlike the two following experiments, frame generation ($\phi^{-1}$) is feasible at this scale and we thus also optimize it in our framework and explicitly sample images at the gradient computation.

\noindent\textbf{Dataset.} 
The moving MNIST dataset generates moving digits by randomly selecting $2$ digits from MNIST dataset \cite{lecun1998gradient} and randomly moving the digits according to the positions in the previous frame. The size of frames are $64 \times 64$. 

\noindent\textbf{Implementation Details.}
We note that the state of this dataset is not fully specified by a single image, because the lack of motion information. We thus define the state as three consecutive images. We extend the architecture of DCGAN~\cite{radford2016unsupervised} to have multi-frame input for frame synthesis, which consists of a 5-layer $3D$ convolutional neural network encoder capturing the information of the image, and a 5-layer $2D$ deconvolutional neural network with upsampling as decoder for image generation (See more detail in supplementary). The encoder is treated as our $\phi(\cdot)$ and decoder is treated as our $\phi(\cdot)^{-1}$. We train on sequences of length $10$. We use curriculum learning to sequentially increase the prediction horizon. At test time, the goal is to generate four consecutive frames given only the initial frame.



\noindent\textbf{Baselines.} 
We compare to the following models:

{\noindent \it  - $L1$ loss.} We use the same network architecture as ours based on DCGAN~\cite{radford2016unsupervised}, but only use the $L1$ reconstruction error of future frame as loss.

{\noindent \it  - LSTM.} We compare to the LSTM prediction model of~\cite{srivastava2015unsupervised}.

{\noindent \it  - Generative Adversarial Network (GAN).} We perform the ablation study of our model by removing the sequence modeling of imitation learning to demonstrate its importance. In this case, the policy directly aims to minimize the local cost function $D$ from the discriminator without sampling it to be its long term expectation $Q$. On the other hand, the discriminator only aims to differentiate a single step generated by our policy from the expert's. This reduces our model to be equivalent to GAN~\cite{goodfellow2014generative}.

\begin{table}
\begin{center}
\begin{tabular}{ |c|c| } 
 \hline
 Accuracy (\%) &  Real sequence \\ \hline
 Real sequence & 50.0 \\ \hline
 Our method & 15.1 \\ \hline
 GAN & 10.2\\ \hline
 LSTM \cite{hochreiter1997long} & 1.4 \\ \hline
 L1 & 0.3 \\ \hline
\end{tabular}
\end{center}\vspace{-2.5mm}
\caption{\small \Hao{Human evaluation} results on the moving MNIST dataset. Our imitation learning framework is able to predict and synthesize more realistic consecutive future frames compared to the baselines because our IRL's minimization of long-term expected cost at training.}
\label{tab:mmnist}
\vspace{-2mm}
\end{table}

\noindent\textbf{Metric.}
We follow previous works~\cite{vondrick2016generating} and evaluate all methods by \Hao{human evaluation (See more detail for the setting of human evaluation in supplementary)}. 
We show on average how many times the output sequence of a method can fool the annotator that it is an actual sequence from the moving MNIST dataset. In this case, the upper bound would be $50\%$ where the generated sequence is the same as real, and the differentiation of real and generated is just by chance.

\noindent\textbf{Results.} 
The result is shown in  \tabref{mmnist}. It verifies that our method can better generate realistic future frames compared to all baselines. The qualitative results are shown in \figref{result_mmnist} (See more qualitative results in supplementary). It is important to note the difference of our method compared to GAN. As shown in \figref{result_mmnist}, the result of GAN becomes much worse and drifts away from the original digits as the time step increases. This is due to the fact that it only considers the local cost at training time in contrast to our IRL framework that handles the compounding error problem in long-term prediction. On the other hand, our method is trained with IRL objective and effectively show high-quality synthesis even for long-term prediction.

\begin{figure}[t]
\begin{center}
  \includegraphics[width=1\linewidth]{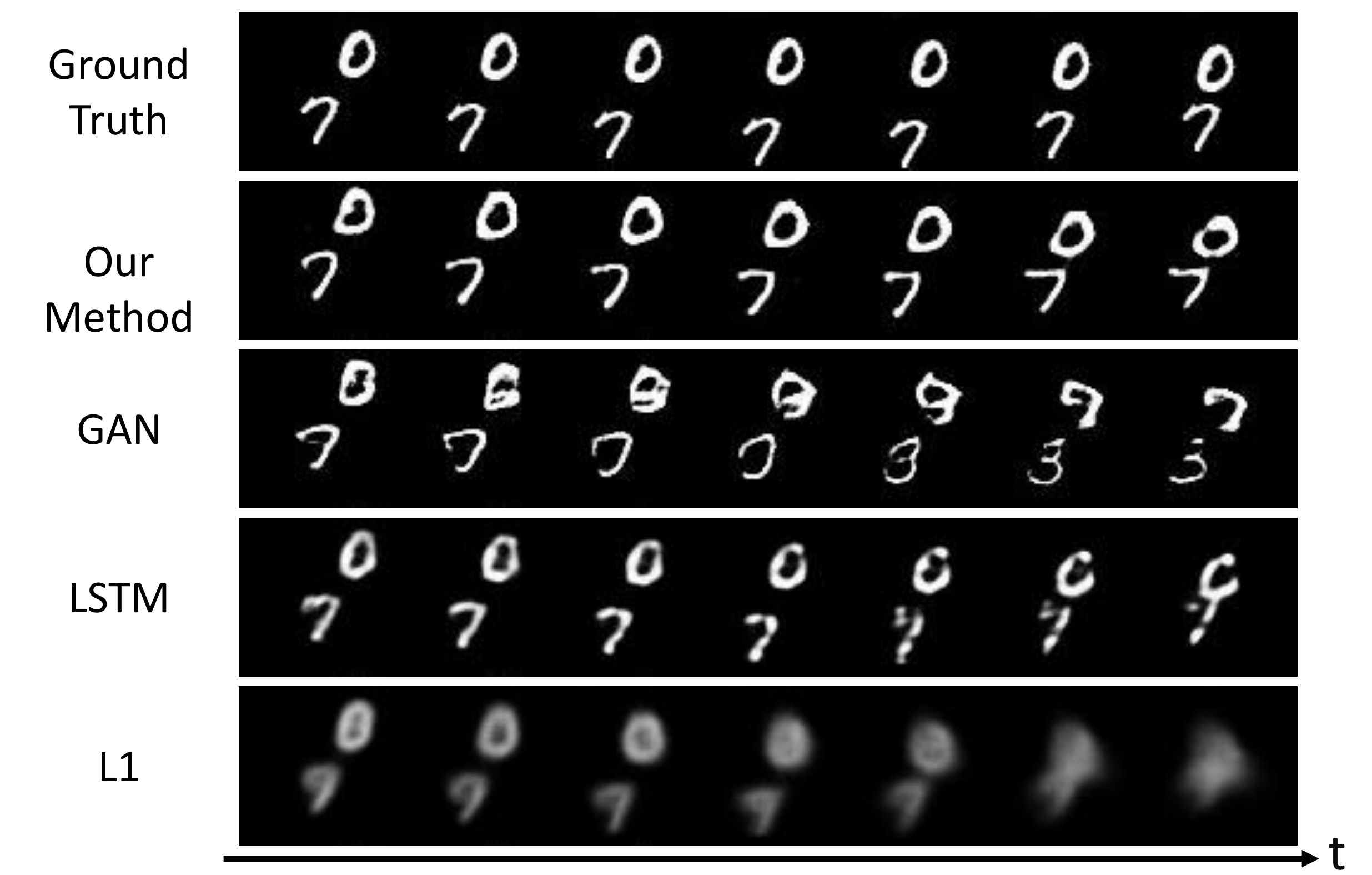}\vspace{-4mm}
\end{center}
   \caption{\small Qualitative results on the moving MNIST dataset. Our model trained with IRL aims to minimize long-term expected cost over the entire sequence. This results in having consistent high-quality frame prediction across time step, and avoids the compounding error problem of baselines.}
   \label{fig:result_mmnist}
   \vspace{-5mm}
\end{figure}

\subsection{Action Prediction on Videos in the Wild}

We examined our method's ability to perform action prediction on the videos in the wild following the setup of~\cite{vondrick2016anticipating}, where the goal is to forecast basic human actions one second before they start. In this experiment,  frames synthesis ($\phi^{-1}$) is no longer computationally feasible at scale. Therefore we need to use our deep reparametrization of the framework in \secref{CNN_IRL}. We demonstrate the model's ability to capture the dynamics of natural videos directly from raw pixels.

\noindent\textbf{Dataset.}  We evaluate our results on TV Human Interactions dataset \cite{patron2010high}, which has 300 videos of four different interactive actions performed by people. We follow previous work~\cite{vondrick2016anticipating} and use the THUMOS15 dataset \cite{THUMOS15} as our unsupervised imitation learning dataset. THUMOS15 consists of trimmed UCF-101 dataset \cite{soomro2012ucf101} and untrimmed background videos. It has $101$ action categories covering a wide range of actions in real world.


\noindent\textbf{Implementation Details.} We follow the training procedure and hyperparameters in \cite{vondrick2016anticipating} and reimplemented their methods for a fair comparison. Our reported number is similar to that reported in~\cite{vondrick2016anticipating}. We train our model with the same setting. We first do imitation learning on the THUMOS dataset with the states/frames being sampled at the same rate as test time (one second in this case). The predictive methods use the ``adapted'' setting in~\cite{vondrick2016anticipating} during training. As it is infeasible to perform frame synthesis in this task, we use our reparametrization of policy. Our $\phi(\cdot)$ uses the architecture in~\cite{he2016deep}. This architecture is also used baselines. Our $\hat{\pi}$ for sampling the learned deep representation uses autoencoder with 3-layer encoder and decoder (See more detail in supplementary).

\noindent\textbf{Baselines.} We compare to the following methods:

{\noindent \it  - SVM.} Two setups on SVM training are considered~\cite{vondrick2016anticipating}. The first setup is to train SVM one second before the action starts in the training time. In this case, the distribution of visual features are the same at both training and testing time. The second is to train SVM during the action at the training time. In this case, the input at testing (one second before the action) has different distribution from training, but training frames are given the correct feature for the action. We refer to the first one as SVM and the second and SVM static. 

{\noindent \it  - LSTM.} We compare to the LSTM prediction model of~\cite{srivastava2015unsupervised} as sequential training baseline that addresses the compounding error problem.

\noindent \textit{- Deep Regression Network}~\cite{vondrick2016anticipating}. We compare to the direct deep feature regression approach of Vondrick \etal~\cite{vondrick2016anticipating}. Their $K=1$ model can be treated as a proxy of ablation study of our approach by replacing IRL with $L2$ regression loss. We also reimplemented their best performing approach with explicit mixture model ($K=3$) for predicting multiple possible futures as reference.

{\noindent \it \Hao{-Pixel Synthesis.}} \Hao{As mentioned in \secref{CNN_IRL}, we resolve the challenge of computational bottleneck from pixel synthesis by introducing a reparametrization that jointly learns a deep representation. It is important to point out that this not only makes our framework efficient, but also makes it effective for semantic prediction. We develop a baseline to demonstrate that our reparametrization is more effective than synthesizing pixels of the future frames. We follow our approach in \secref{FFG} and apply the model to downsampled version (64 x 64) of the videos. We anticipate the action with the same setting used by other baselines. The only difference is that now the SVM is applied to deep features extracted from the synthesized future frames rather than the deep representation directly predicted by our model.}

\begin{figure}[t]
\begin{center}
  \includegraphics[width=1\linewidth]{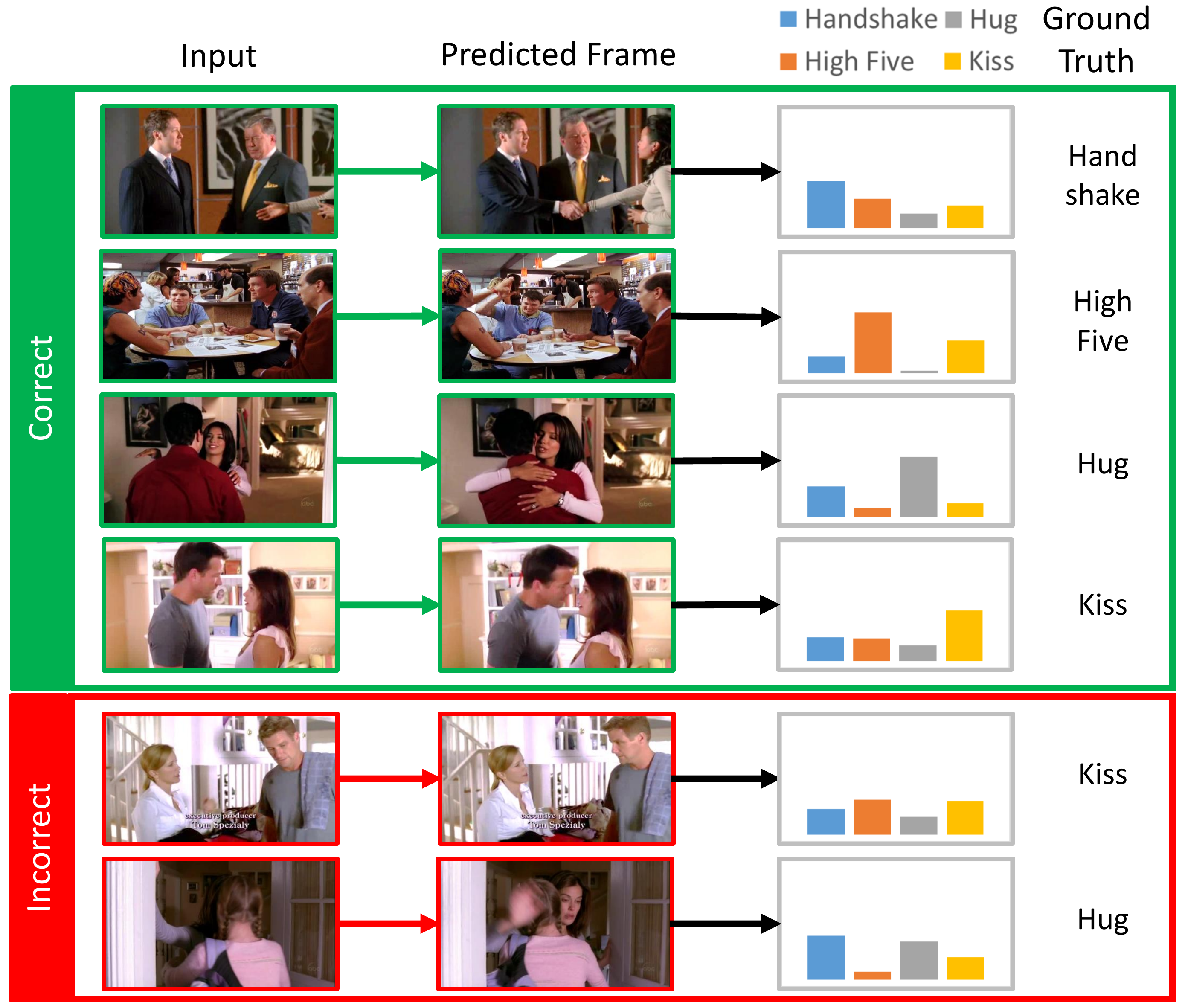}\vspace{-4mm}
\end{center}
   \caption{\small
   Single step action prediction results. Given an input frame one second before the the interaction takes place, our model is able to robustly predict the future frame and improve significantly over the baselines for action prediction. The frame visualization is retrieved by nearest neighbor image of the predicted deep representation.
   }
   \label{fig:result_action}
   \vspace{-2mm}
\end{figure}

\begin{table}
\begin{center}
\begin{tabular}{ |c|c| } 
 \hline
 Model &  Accuracy (\%) \\ \hline
 Random & 25.0 \\ \hline\hline
 SVM static & 36.2 \\ \hline
 SVM & 35.8 \\ \hline\hline
 Deep Regression ($K=1$)~\cite{vondrick2016anticipating} & 37.5 \\ \hline
  LSTM~\cite{srivastava2015unsupervised} & 40.5 \\ \hline
 \Hao{Pixel Synthesis} & 26.1 \\ \hline
 Ours & \textbf{45.8} \\ \hline \hline
 Deep Regression ($K=3$)~\cite{vondrick2016anticipating} & 44.0 \\
 \hline
\end{tabular}
\end{center}\vspace{-2.5mm}
\caption{\small Accuracy of action prediction on the TV Human Interaction dataset. Our learned policy significantly outperforms reconstruction loss based baselines. Our single mode policy also outperforms deep mixture regression model which is designed specifically for this task.}
\label{tab:action}
\vspace{-5mm}
\end{table}

\begin{figure}[t]
\begin{center}
  \includegraphics[width=1\linewidth]{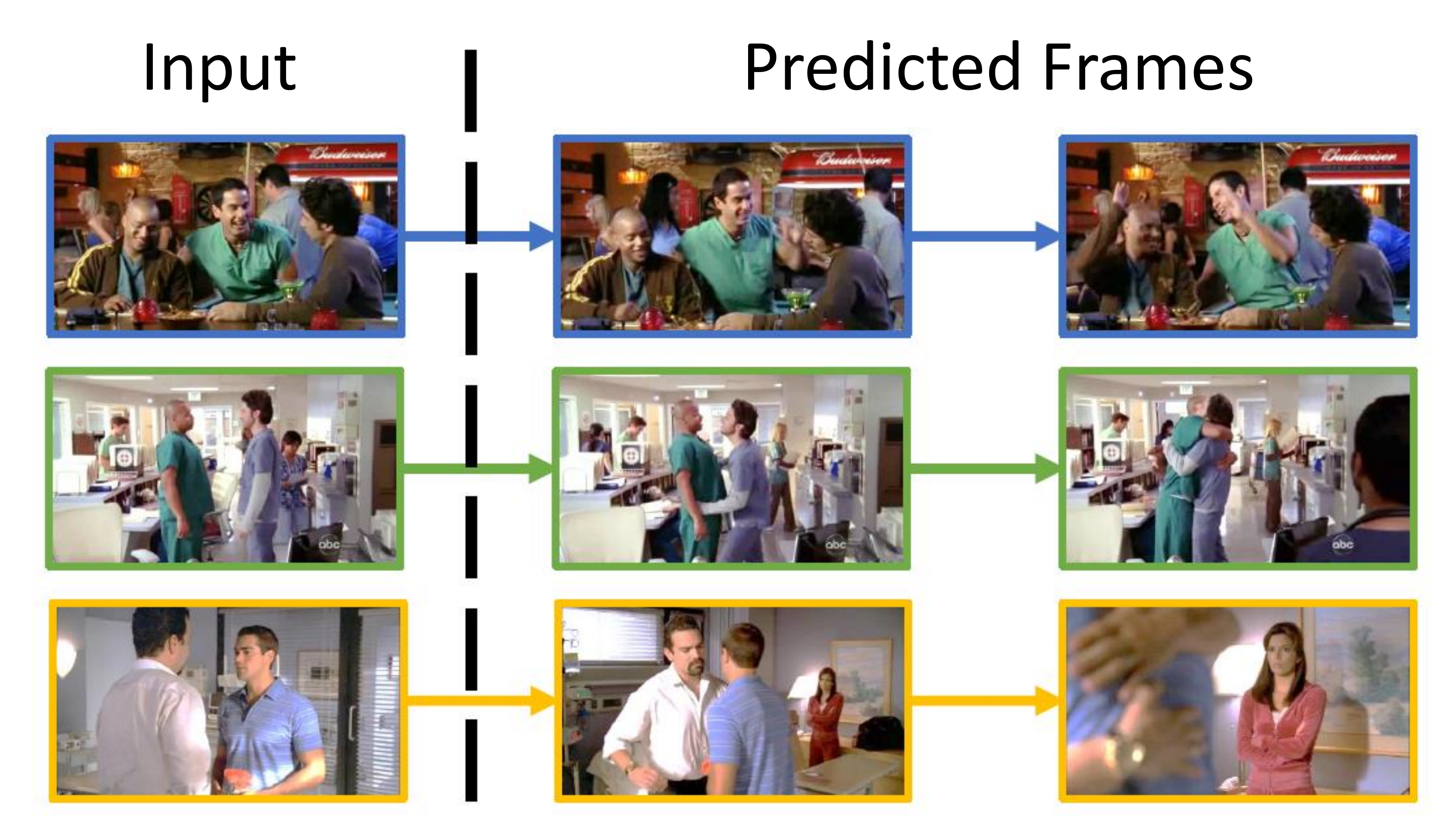}\vspace{-4mm}
\end{center}
   \caption{\small
   Multi-step video prediction results. Given only a single frame as input, our learned policy is able to accurately forecast multiple future frames in natural video. This verifies the effectiveness of our IRL formulation and reparametrization that allows direct imitation from the pixel level. Images are retrieved by nearest neighbor of predicted deep feature.
   }
   \label{fig:result_action_long}
   \vspace{-5mm}
\end{figure}

\noindent\textbf{Results.} The results for action anticipation are shown in \tabref{action}. Our IRL formulation of action prediction significantly outperforms the corresponding version using just $L2$ loss (Deep Regression $K=1$). In addition, we are able to outperform the sequential baseline using just the reconstruction loss (LSTM). It is also important to note that our single mode policy without mixture model is already able to outperform the best performing Deep Regression with explicit mixture model for handling multiple possible futures. Note that our reported reimplementation of $K=1$ is slightly worse than their reporting number because we perform unsupervised learning for all methods only on the THUMOS dataset for a fair comparison. This indicates that single mode regression is more prone to be affected by dataset bias and demonstrate the robustness of our method. \Hao{On the other hand, the accuracy of Pixel Synthesis is $26.1\%$, which is lower than our framework by a large margin. This verifies the importance of bypassing pixel synthesis in our framework for anticipating action in natural videos.} Qualitative results of action prediction is shown in \figref{result_action} (See more qualitative results in supplementary). It can be seen that our learned policy robustly predicts the action after one second. We further demonstrate that our model is able to predict more than one step into future frames. The qualitative results are shown in \figref{result_action_long}. This shows the effectiveness of our imitation learning framework for visual prediction, and verifies that our model can imitate natural video from the pixel value using our reparametrization of deep representation.



\subsection{Storyline Forecasting on Semantic Dynamics}

In the final experiment, we evaluate our method on the challenging semantic forecasting in visual storylines~\cite{huang2016visual,sigurdsson2016learning}. Given a storyline consisting of images, our goal is to predict the next image. This is challenging as the change of semantic meaning might not be capture directly through visual feature matching. Our setup is similar to the prediction task of~\cite{sigurdsson2016learning}.

\begin{figure}[t]
\begin{center}
  \includegraphics[width=1\linewidth]{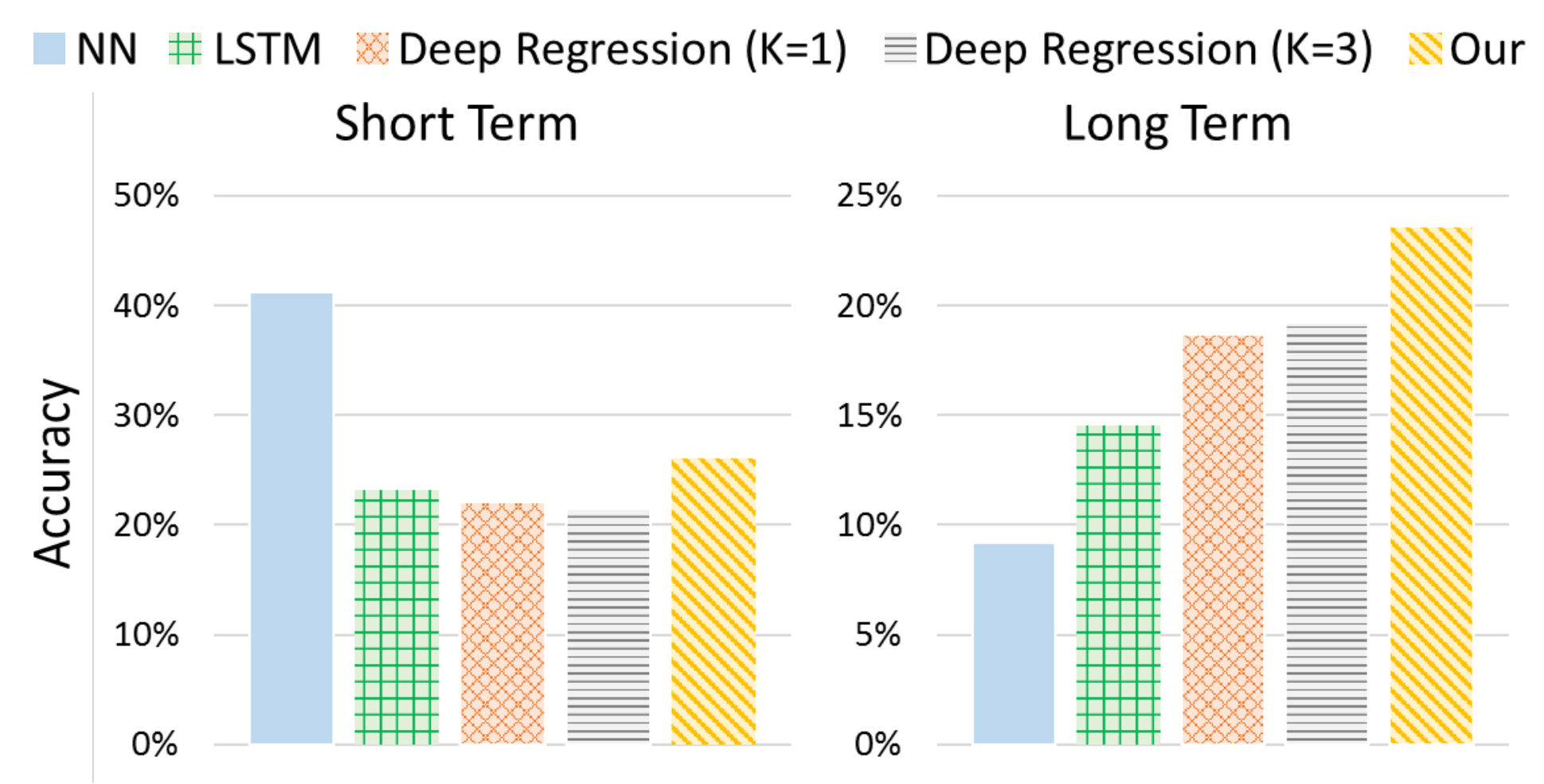}
\end{center}\vspace{-4mm}
   \caption{\small
   Accuracy of storyline prediction on the Visual Storytelling Dataset. We follow previous work and evaluate short-term and long-term prediction separately. It can be seen that nearest neighbor on the deep feature representation is a strong baseline is this case because the goal is similar to retrieve visually similar frames. On the other hand, nearest neighbor does not work for long-term prediction. Our general visual prediction framework performs the best for the long-term prediction. This verifies that our model can also be applied to the challenging semantic  prediction, while learning directly from the pixel.
   }
   \label{fig:result_storyline}
   \vspace{-5mm}
\end{figure}

\noindent\textbf{Dataset.} We use the Visual Storytelling Dataset (VIST)~\cite{huang2016visual} for our storyline forecasting task. The dataset consists of photos from more than 100,000 albums from Flickr. We randomly select one photo sequence of storyline from each album. Using the original dataset split in the dataset, this results in total 8024 training, 1011 testing, and 998 validation storylines, where each consists of five photos illustrating a story. Note that while we follow the evaluation setup of~\cite{sigurdsson2016learning}, their dataset is not applicable to our task as they only have 10 storylines per concept for evaluation only and does not have the large scale human demonstrated storyline available for imitation learning.

\noindent\textbf{Experimental Setup and Metrics.} We follow~\cite{sigurdsson2016learning} and split the storyline evaluation into two goals: the first is short-term, where the image is visually similar consecutive image in the album. The second goal is long-term prediction, where prediction is for the next representative event involving large visual appearance change. We follow~\cite{sigurdsson2016learning} and posed this as a classification task, where the goal is to select the true image along with other four other images selected randomly from the same album.  We use the training set storylines as expert demonstrations for both our imitation learning and unsupervised learning of baselines. We report resulting accuracy for the test split on both the short-term and long-term prediction (See more detail for experimental setup in supplementary). 

\noindent\textbf{Baselines.} We again compare to LSTM~\cite{srivastava2015unsupervised} and Deep Regression Network~\cite{vondrick2016anticipating} from the previous task. In addition, we compare to \emph{nearest neighbor} of the ResNet-101 feature representation~\cite{he2016deep}, which is proven to be a strong baseline for short-term prediction~\cite{sigurdsson2016learning}. 

\noindent\textbf{Results.} The results of short-term and long-term prediction are shown in \figref{result_storyline}. For short term prediction, we observe the similar results as with in~\cite{sigurdsson2016learning}, where nearest neighbor is the strongest baseline as the goal is similar to finding the photo with the most similar visual feature. It is important to note that our method still outperforms reconstruction loss based baselines (LSTM and Deep Regression Network) in this case. For the long-term prediction, it can be seen that the nearest neighbor of visual appearance matching no longer works and performs poorly. On the other hand, our direct imitation of storyline photos from the pixel values is proven effective compared to both LSTM and Deep Regression Networks. This verifies that our approach is general and can even applied to imitate human behavior at the challenging semantic-level. Qualitative results are shown in \figref{storyline} (See more qualitative results in supplementary). It can be seen from the figure that our method can better predict semantically meaning next photo in the storyline. First three rows show our correct prediction, and the final row shows incorrect prediction. It can be seen that our method is able to capture the semantic meaning (event or location) of the input photo and continue to forecast photos with similar semantic meaning. 

\begin{figure}[t]
\begin{center}
  \includegraphics[width=1\linewidth]{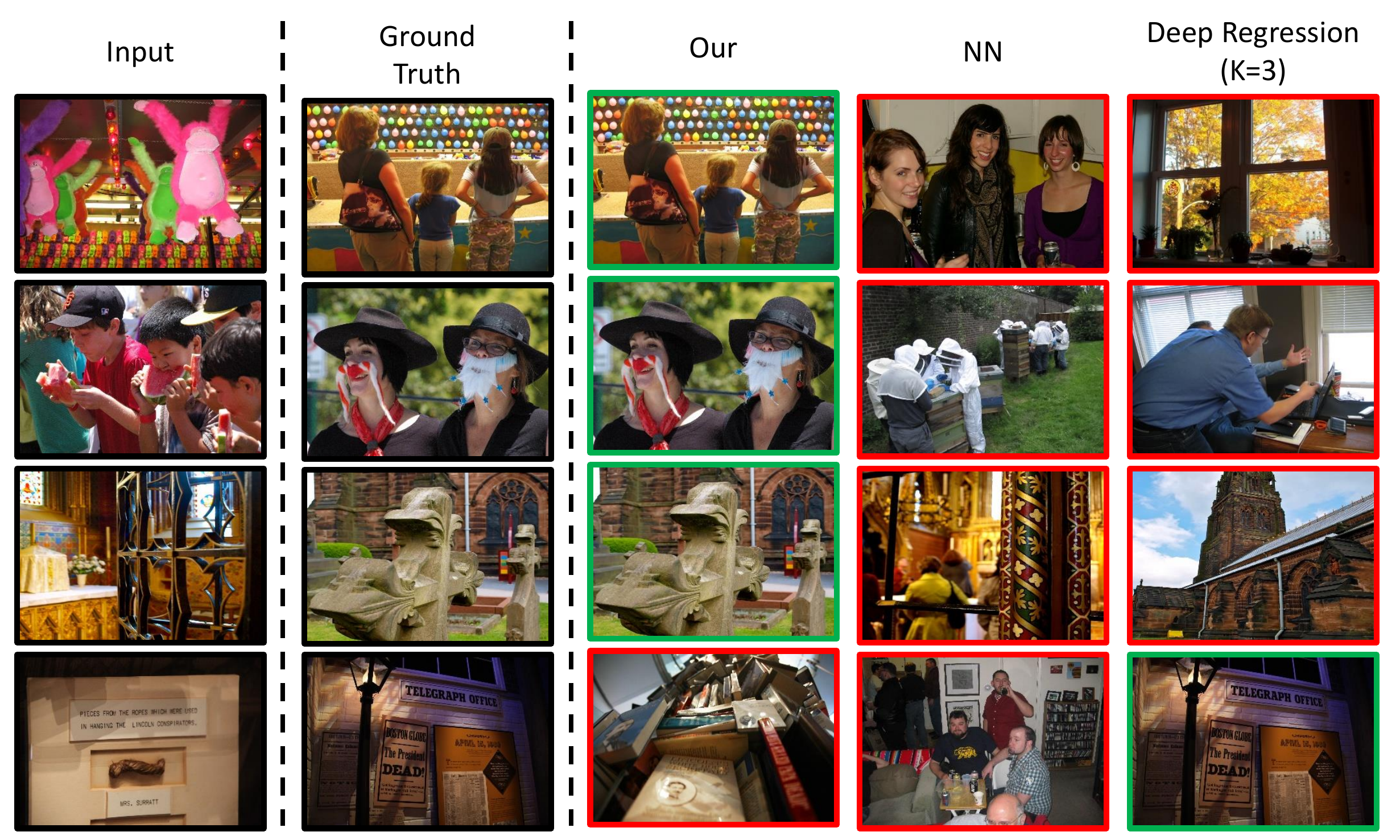}\vspace{-4mm}
\end{center}
   \caption{\small Qualitative results of storyline forecasting. Green boxes indicate correct prediction of the next image in the storyline, and red boxes indicate incorrect predictions.   
   Our imitation learning based framework can best capture the semantics by imitating storylines generated by human. On the other hand, baselines focusing on appearance matching or reconstruction loss is harder to successfully predict semantics in the storylines.}
   \label{fig:storyline}
   \vspace{-5mm}
\end{figure}



\section{Conclusion}\label{Con}
We have presented a general framework for long-term visual prediction. We formulate this as an IRL problem, and extend previous works to directly imitate natural visual sequences from the pixel level. The key challenge is the high-dimensional and continuous natural of images. We address this by (1) introducing dual formulation of IRL to natural visual sequences that bypasses exhaustive state-action pair visit, and (2) reparametrization using deep representation to avoid explicit state synthesis during gradient computation. We demonstrated that our framework is general and effective using three tasks at different level of abstractions: (1) future frame generation, (2) action prediction, (3) storyline forecasting. We verified that our unified framework is able to outperform existing methods on all three tasks.






\vspace{5mm}
\noindent\textbf{Acknowledgement.} We thank National Taiwan University (NTU-106R104045), NOVATEK Fellowship, MediaTek, Intel, Naval Research (N00014-15- 1-2813), and Panasonic for their support. We thank Danfei Xu, Linxi Fan, Animesh Garg, Tseng-Hung Chen and Hou-Ning Hu for helpful comments and discussion.



\end{document}